\documentclass{opt2026}

\newcommand{\changescolour}{blue}

\newcommand{\matern}{Mat\'ern}

\title[Anisotropic Trust Region Warping for PFNs in Bayesian Optimisation]{Anisotropic Trust Region Warping for Prior-Data Fitted Networks in Bayesian Optimisation}

\begin{document}

\makeatletter
\g@addto@macro\normalsize{%
  \setlength{\abovedisplayskip}{5pt}%
  \setlength{\belowdisplayskip}{5pt}%
  \setlength{\abovedisplayshortskip}{2pt}%
  \setlength{\belowdisplayshortskip}{3pt}%
}
\makeatother
\setlength{\textfloatsep}{6pt}
\setlength{\intextsep}{8pt}

\optauthor{%
\Name{Samuel Younger} \Email{samuel.younger24@imperial.ac.uk}\\
\addr Imperial College London
\AND
\Name{Tinkle Chugh} \Email{t.chugh@exeter.ac.uk}\\
\addr University of Exeter
\AND
\Name{George {De Ath}} \Email{g.de.ath@exeter.ac.uk}\\
\addr University of Exeter%
}

\maketitle
\vspace{-15pt}
\begin{abstract}%
Prior-Data Fitted Networks (PFNs) replace the hyperparameter refitting of Gaussian Process (GP) surrogates in Bayesian optimisation with a single forward pass, at the cost of a prior that is fixed at meta-training time. We show empirically that a PFN's predictive log-likelihood falls increasingly behind that of a refitted GP as the objective function's lengthscale grows beyond the mode of the PFN's meta-training prior, and that this deficit carries over to optimisation performance. To correct the mismatch without retraining the network, we introduce Anisotropic Trust Region (ATR) warping. Per-dimension lengthscales are estimated by kernel target alignment, and an affine rescaling of the input space maps the objective onto the lengthscale the PFN expects, with a trust region around the anchor determining which observations are retained, leaving the PFN's $\mathcal{O}(d \cdot N^2)$ inference cost unchanged. On functions drawn from a \matern-3/2 kernel, ATR warping improves optimisation performance of the PFN at both extremes of the lengthscale range. Our implementation is publicly available at https://github.com/youngers2006/Trust-Region-Warping-For-PFNs.
\end{abstract}

\section{Introduction}
Bayesian Optimisation (BO)~\citep{garnett2023bayesian} has traditionally relied on Gaussian Processes (GPs) as surrogate models due to their rigorous uncertainty quantification and exact posterior inference. However, GPs impose a severe computational bottleneck, refitting the covariance matrix requires $\mathcal{O}(N^3)$~\citep{rasmussen2006gaussian} complexity, limiting their scalability for optimisation tasks.

Prior-Data Fitted Networks (PFNs) have emerged as a highly scalable alternative to GPs~\citep{muller2023pfns4bo}. By training Transformers to approximate Bayesian posteriors, PFNs perform inference entirely via in-context learning~\citep{Muller2021}, reducing the cost of refitting to $\mathcal{O}(1)$ and bounding inference to $\mathcal{O}(d \cdot N^2)$.

Despite these computational advantages, this research highlights an issue in PFNs, namely poor generalisation to objective functions with lengthscales above the mode of its meta-training prior. Because PFNs are amortised surrogates, their predictive accuracy is bound to the static prior they were trained on. Through stress testing via prior-objective mismatch, we demonstrate that when evaluating functions with lengthscales well above the mode of the PFN’s training prior, the model’s predictive log-likelihood and downstream optimisation performance diverge from that of a GP. 

To solve this degradation without sacrificing scalability, we introduce Anisotropic Trust Region (ATR) Warping. Rather than expanding the PFN's training distribution, ATR warps the input space to match the PFN’s native prior. By optimising a target lengthscale vector using Kernel Target Alignment and constraining the search space, we map mismatched functions back towards the mode of the PFN's prior. We then show that this method improves the performance of the PFN on objective functions with mismatched lengthscales whilst preserving its $\mathcal{O}(d \cdot N^2)$ complexity. \looseness=-1

\section{Background and Motivation}
\label{sec:background}
PFNs amortise Bayesian inference via meta training, replacing the hyperparameter fitting done in GPs with a forward pass~\citep{Muller2021}. Amortisation limits the model's ability to fit the target functions that lie outside of the meta-training prior. A PFN is optimal only under the prior it was meta-trained on~\citep{nagler2023statistical}, and its in-context predictions are not known to generalise beyond that prior's support~\citep{muller2024bayes}.\looseness=-1

To demonstrate this effect we tested the model fitting capability of the PFN against the GP across a range of lengthscales. We draw functions from a \matern-3/2 kernel, using random Fourier feature sampling~\citep{rahimi2007random}, of unit amplitude with lengthscale, $l$, swept over $[0.021, 2.0]$, and measure the predictive log-likelihood of the objective function averaged over test locations, given $20 \cdot d$ observations, reporting the median across function draws. As a reference we fit a GP with the same kernel and ARD, refitted by marginal likelihood at every step under a deliberately uninformative smoothed box prior which includes the lengthscales tested. A reference is necessary because absolute log-likelihood rises with $l$ for both models. Longer lengthscales mean more effective data per correlation length, so the task itself becomes easier, and only the relative comparison isolates the fixed prior.\looseness=-1
\begin{figure}[htbp]
\centering
\includegraphics[width=0.5\textwidth]{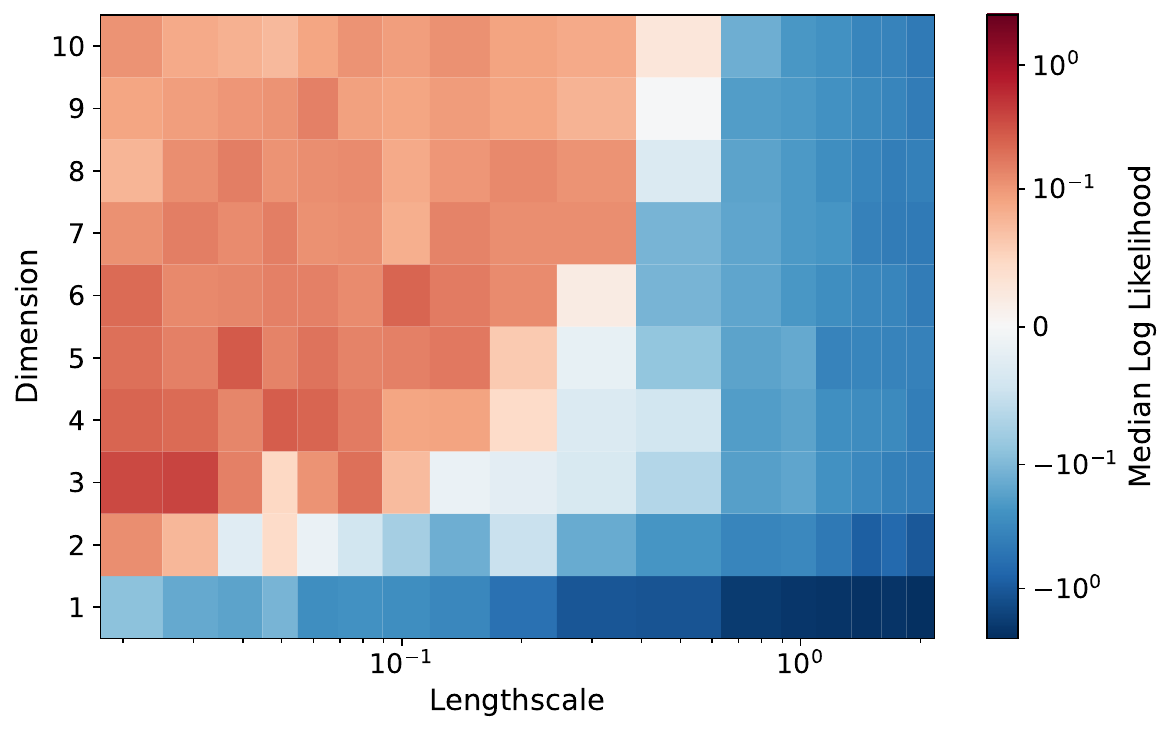}
\vspace{-11pt}
\caption{Median predictive log-likelihood of the PFN minus that of a refitted GP. Positive values (red) mean the PFN fits better, negative (blue) the GP; the colour bar is symmetric log.}
\label{lengthscale_heatmap}
\end{figure}

Figure~\ref{lengthscale_heatmap} shows the difference in median log-likelihood across seventeen lengthscales and ten dimensions. In every dimension tested the PFN falls progressively further behind the GP as the lengthscale grows, and the boundary at which the two models demonstrate equal performance moves to larger lengthscales as dimension increases. We attribute this to sampling density rather than any change in the underlying mechanism, under a budget of $20 \cdot d$, observations per axis scale as $(20 \cdot d)^{\frac{1}{d}}$, so data per correlation length drops sharply with dimensionality, the same effect that motivates dimension-scaled lengthscale priors~\citep{hvarfner2024vanilla}. Because a PFN's context is bounded to a few hundred points~\citep{muller2023pfns4bo}, sampling density cannot be held constant across dimensions.\looseness=-1
\begin{figure}[htbp]
\centering
\includegraphics[width=0.6\textwidth, trim={0 7pt 0 7pt}, clip]{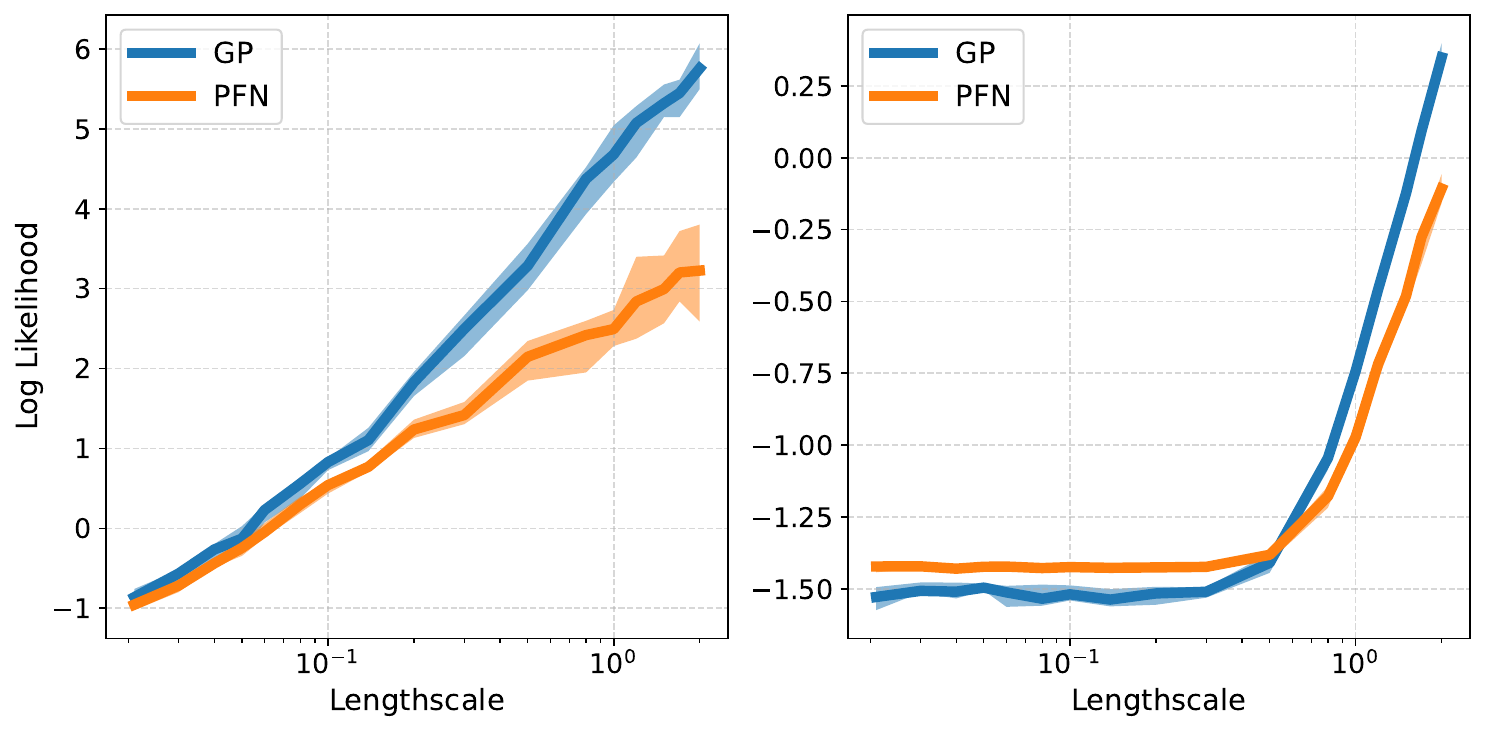}
\vspace{-11pt}
\caption{1D and 10D regression performance of the GP and PFN across a range of lengthscales.}
\label{ls_linegraphs}
\end{figure}

The colour scale in Figure~\ref{lengthscale_heatmap} compresses the higher dimensions, where the deficit is real but smaller, so Figure~\ref{ls_linegraphs} shows the underlying curves at $d = 1$ and $d = 10$. In one dimension the two models are indistinguishable up to approximately $l = 0.1$, after which the PFN performance degrades steadily, reaching a deficit of roughly 2.5 nats at $l = 2.0$. The onset of divergence coincides with the PFN's nominal prior mode, $l = 0.1385$~\citep{muller2023pfns4bo}, even though these lengthscales lie within the bulk of the prior (its mean is $0.80$), so the degradation tracks distance from the prior's mode rather than departure from its support. Below the onset both models sit close to their predictive priors, neither is learning, and the small residual favouring the PFN reflects that a meta-learned prior is informative where an uninformative box prior is not. At $d = 10$ the same shape appears with the onset shifted to approximately $l = 0.5$ and a deficit of roughly 0.4 nats.

The deficit is also visible in optimisation performance, though not uniformly: in 2D the gap in normalised simple regret between GP and PFN peaks at intermediate lengthscales and narrows at both extremes, since the cost of a posterior error depends on how much the acquisition decision turns on it. At $d = 10$ only the GP makes progress, and only at the longer lengthscales. Full regret curves are given in Appendix~\ref{app:regret}.

A GP addresses lengthscale mismatch by fitting the lengthscale. A PFN cannot, but the same correction can be applied in the input space: rescaling the domain by a factor $s$ divides the function's lengthscale by $s$, so a suitable choice of $s$ maps a mismatched function onto one that the prior expects. \citet{snoek2014input} warped the inputs in this way so that a stationary GP could fit non-stationary objectives, and LABCAT~\citep{visser2025labcat} rescales its trust regions using the lengthscales of a local GP, but both rely on a surrogate that is refitted as data arrives. TuneTables~\citep{feuer2024tunetables} and GIT-BO~\citep{yu2025gitbo} do adapt a frozen PFN, though through learned context embeddings and a gradient-informed search subspace respectively, rather than through the inputs, and we are not aware of work that warps the inputs to match a fixed PFN's prior. Without refitting, the warp needs its own estimate of the lengthscale, so prior mismatch becomes an estimation problem. Section~\ref{sec:method} describes our estimator and the warp built from it, and Section~\ref{sec:experiments} evaluates the result.

\section{Anisotropic Trust Region Warping}
\label{sec:method}
\textbf{Aim:} Estimate the objective function's lengthscale on $[0,1]^d$ and use it to construct an anisotropic trust region, $\mathcal{T}$, whose affine warp maps the lengthscale to the PFN prior's mode, in $\mathcal{O}(d \cdot N^2)$~\citep{Muller2021}.

\textbf{Estimating the Lengthscale:}
Kernel Target Alignment (KTA)~\citep{cristianini2001kernel} estimates the lengthscale from sparse samples by aligning a kernel from the PFN's known prior family to the data, and scales as $\mathcal{O}(d \cdot N^2)$. To use KTA we optimise a lengthscale vector $l$ with gradient descent to maximise the normalised Frobenius inner product between a target matrix $K_y = y y^T$, where $y$ is centred around its mean~\citep{cortes2012algorithms}, and a spatial \matern-3/2 kernel matrix $K_x$:
\begin{equation}
A(l) = \frac{\langle K_x, K_y \rangle_F}{\|K_x\|_F \, \|K_y\|_F}.
\end{equation}
Appendix~\ref{app:kta} reports the accuracy of this estimator across the tested range.

\textbf{Trust Region Sizing:}
With the estimated lengthscale, $l$, we construct an anisotropic trust region centred at the current anchor, $x_{\mathrm{opt}}$. TuRBO~\citep{eriksson2019turbo} also shapes its trust regions anisotropically, with each side proportional to a fitted GP lengthscale, but there the lengthscales set only the region's aspect ratio and its volume is adapted by success and failure counters. Here the lengthscales set the side lengths directly. The warping factor $s_j$ for each dimension $j$ is set by the ratio of the estimated lengthscale $l_j$ to the PFN's target lengthscale $l_{\mathrm{target}}$, and the side length $L_j$ follows from it:
\begin{equation}
s_j = \max \left( s_{\mathrm{min},j}, \frac{l_j}{l_{\mathrm{target}}} \right),
\qquad
L_j = \min \left( 1, s_j \right),
\qquad
u_j = \min \left( 1, \frac{1}{s_j} \right).
\end{equation}

The warping factor $s_j$ is the divisor of the affine mapping, while the window size $L_j$ bounds the region of the domain in which the objective function can be queried, and cannot exceed 1. When $l_j < l_{\mathrm{target}}$ the objective is rougher than the prior expects, $s_j < 1$, and a sub-region of size $s_j$ is warped to $[0,1]$. Warping the full domain by the same factor would map it onto $\Pi_j [0, 1/s_j]$, which we expect to be outside the unit cube on which the PFN was meta-trained when $l_j < l_{target}$. The trust region is therefore what makes the warping admissible for lengthscales below the mode. When $l_j > l_{\mathrm{target}}$ the objective is smoother than expected, $s_j > 1$, and the region of side $s_j$ extends beyond the domain, so all data is retained ($L_j = 1$). Warping that region to $[0,1]$ places the real domain in $[0, u_j]$, the remainder cannot be queried.\looseness=-1

\textbf{Trust Region Placement:}
The region is centred on $x_{\mathrm{opt}}$ with side lengths $L_j$, and is translated back inside the domain rather than truncated where it would overflow $[0,1]$, so its width is preserved.

\textbf{Restarts:}
After $\tau$ consecutive iterations without improvement, as in TuRBO, a point is drawn uniformly from the domain and becomes the new anchor. Restarts are necessary at smaller lengthscales, where rough functions have many basins, to prevent getting stuck in local optima.

\textbf{Bounding Against Context Starvation:}
Highly rough functions can shrink the region so far that it discards past evaluations and leaves the surrogate with no context. To prevent this, the floor is set from the $k$ observations nearest $x_{\mathrm{opt}}$ in the $L_\infty$ (Chebyshev) norm, with axis $j$ scaled by $l_j / l_{\mathrm{target}}$ so that the neighbourhood reflects the anisotropy the warp will apply. Writing $d_{k,j}$ for their largest deviation from $x_{\mathrm{opt}}$ in dimension $j$, we set $s_{\mathrm{min},j} = 2 d_{k,j}$, the axis-aligned bounding box of those $k$ points, which ensures the region contains at least $k$ samples.

\textbf{Affine Warping and Inference:}
Once a trust region with bounds $[a_j, b_j]$ for $j = 1, \dots, d$ has been defined, the data is filtered to isolate only the samples $x_i$ located inside the hypercube $\mathcal{T}$:
\begin{equation}
    x_i \in \mathcal{T} \iff \bigwedge_{j=1}^d (x_{ij} \in [a_j, b_j]).
\end{equation}

An affine transformation can then be applied such that $[a_j , b_j] \rightarrow [0 , u_j]$ in each dimension:
\begin{equation}
    \psi_j(x_j) = \frac{x_j - a_j}{s_j}.
\end{equation}

The hypercube $\mathcal{T}$ contains a subset of the sampled dataset that has been warped to match the lengthscale of the PFN training prior. This subset is passed to the PFN for a forward pass, and the acquisition function is optimised over $\Pi_j [0, u_j]$. Algorithm~\ref{alg:atr} summarises the procedure.

\section{Experiments and Results}
\label{sec:experiments}
\begin{figure}[t]
\centering
\includegraphics[width=0.95\textwidth, trim={0 7pt 0 7pt}, clip]{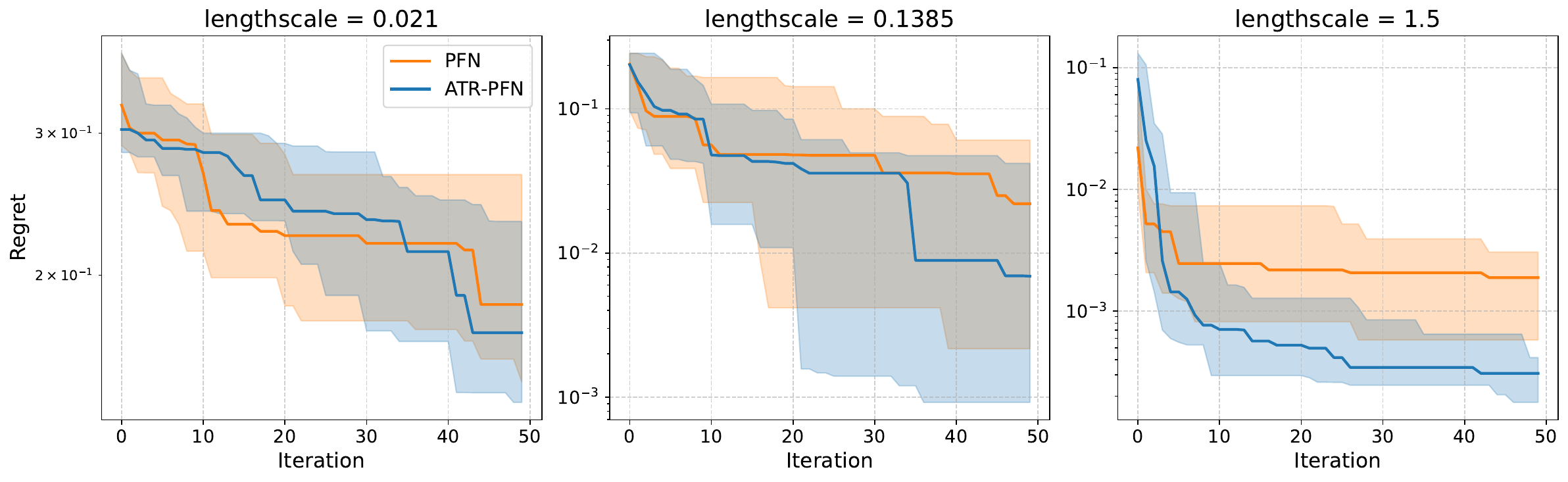}
\vspace{-11pt}
\caption{2D normalised regret curves (median and IQR over 21 repeats) for PFN and ATR-PFN. The centre panel is the prior mode ($l=0.1385$), the outer panels the tested extremes.}
\label{2D_regret_plot}
\end{figure}

To demonstrate the effectiveness of ATR warping we tested it against the base PFN on functions drawn from a \matern-3/2 kernel across 15 lengthscales in $[0.021, 1.5]$, at $d \in \{ 2,5,10 \}$, running 50 iterations with 21 repeats per configuration. Draws are isotropic so that one lengthscale indexes the sweep; anisotropic draws gave the same picture. Figure~\ref{2D_regret_plot} shows the 2D results at both extremes and at the prior mode; Appendix~\ref{app:atr-regret} has the full set and Appendix~\ref{app:Exp_det} the setup.

ATR warping is beneficial at extreme lengthscales, though the improvement is more modest at the lower end. For lengthscales in distribution the estimated ratio $l_j / l_{\mathrm{target}}$ is close to one, so the trust region covers most of the domain, and ATR-PFN behaves much like the PFN. What improvement remains comes from the trust region and restarts rather than warping, which is near identity here. The same pattern holds at 5D and 10D (Figures \ref{5D_sweep_plot_comp} and \ref{10D_sweep_plot_comp}), though, as Figure~\ref{lengthscale_heatmap} would predict, with smaller margins. An ablation in Appendix~\ref{app:ablation}, supplying the true lengthscale in place of the KTA estimate, shows that estimator variance does not drive the improvement. At a larger budget of 150 iterations the advantage for smaller lengthscales widens substantially (Figure~\ref{2D_sweep_plot_comp_150}), we do not separate the contributions of the warping and the trust region at short lengthscales.

\section{Discussion and Future Work}

We have shown that a fixed PFN's predictive accuracy and downstream optimisation performance degrade as the objective's lengthscale grows beyond the mode of its meta-training prior, and that ATR warping substantially improves the PFN at mismatched lengthscales, without retraining the network or raising its $\mathcal{O}(d \cdot N^2)$ inference complexity. The ablation in Appendix~\ref{app:ablation}, which supplies the true lengthscale in place of an estimate, suggests the improvement is not driven by estimator variance. Above the prior mode $L_j = 1$, so the region spans the domain and restarts cannot move it, neither localisation mechanism is active there, and the gain that remains is attributable to the warping alone.

KTA's estimates are monotone but shrunk toward the centre of the range (Appendix~\ref{app:kta}), and have high variance on sparse data, which carries through to the run-to-run variance of the optimisation; a better low-sample estimator would reduce both. High alignment is also only a sufficient, not a necessary, sign of a good kernel~\citep{nguyen2008efficient}. Shrinking the region to correct a rough function discards evaluations outside it, so the k-nearest-neighbour floor keeps at least $k$ points in context, at the cost of capping the warp. Context and correction therefore pull against each other, and in high dimensions, where the $k$ nearest evaluations are far apart, the floor keeps the region large and the warp has little effect.\looseness=-1

We isolated a single axis of prior mismatch, and how the method behaves when several are misspecified at once remains to be tested on standard benchmarks. ATR leaves the network untouched, so it composes with corrections for other axes, such as HEBO's Kumaraswamy input warping~\citep{CowenRivers2020}.

\bibliography{sample}

\newpage
\appendix

\section{Regret Curves - GP Comparison}
\label{app:regret}

\begin{figure}[htbp]
\centering
\includegraphics[width=0.95\textwidth]{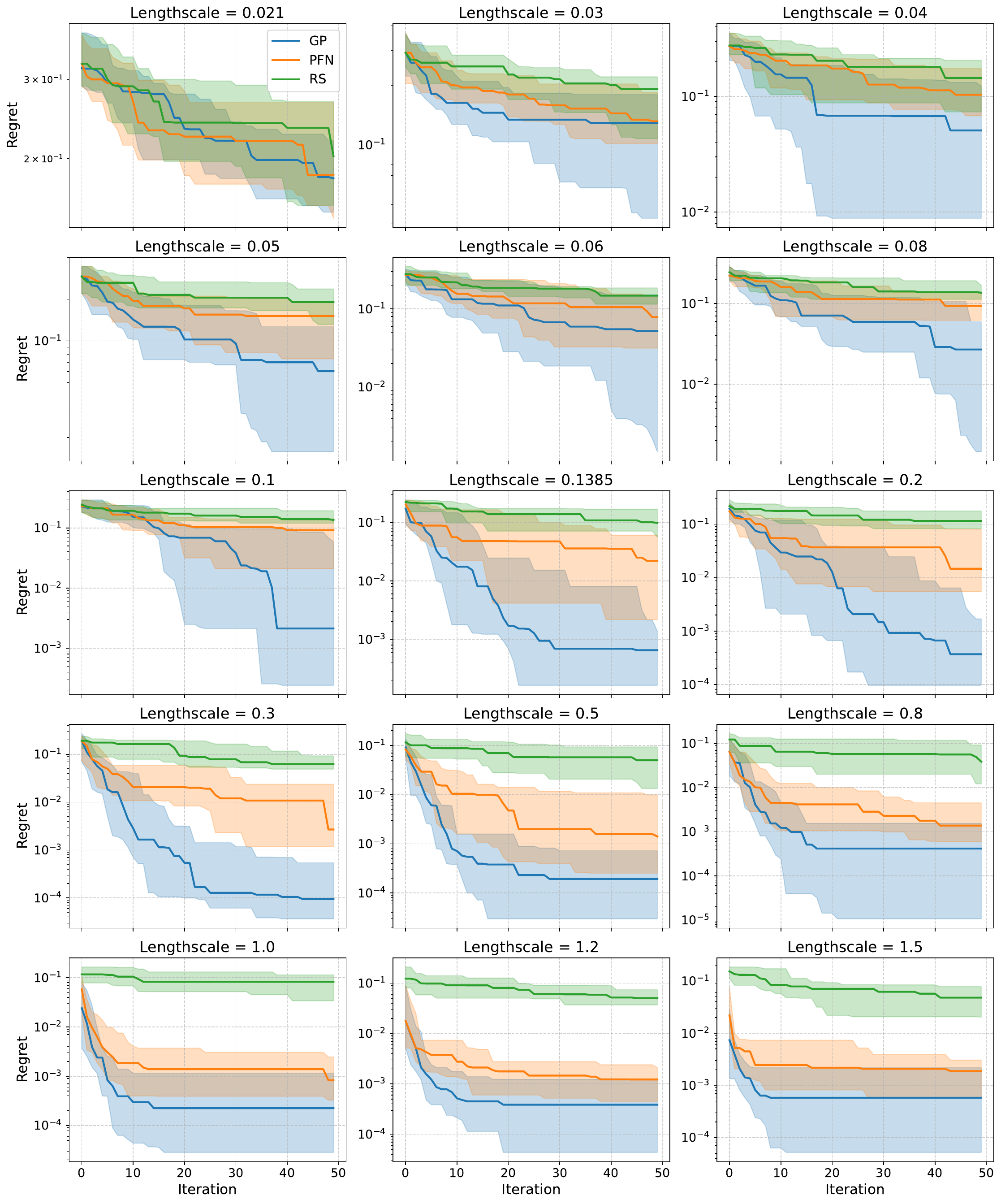}
\caption{2D normalised regret curves (median and IQR over 21 repeats) for GP, PFN and random search across 15 lengthscales. The PFN's prior mode is $l=0.1385$, panels to the left of it are increasingly rough relative to the prior, and those to the right increasingly smooth.}
\label{2D_sweep_plot}
\end{figure}

\begin{figure}[htbp]
\centering
\includegraphics[width=0.95\textwidth]{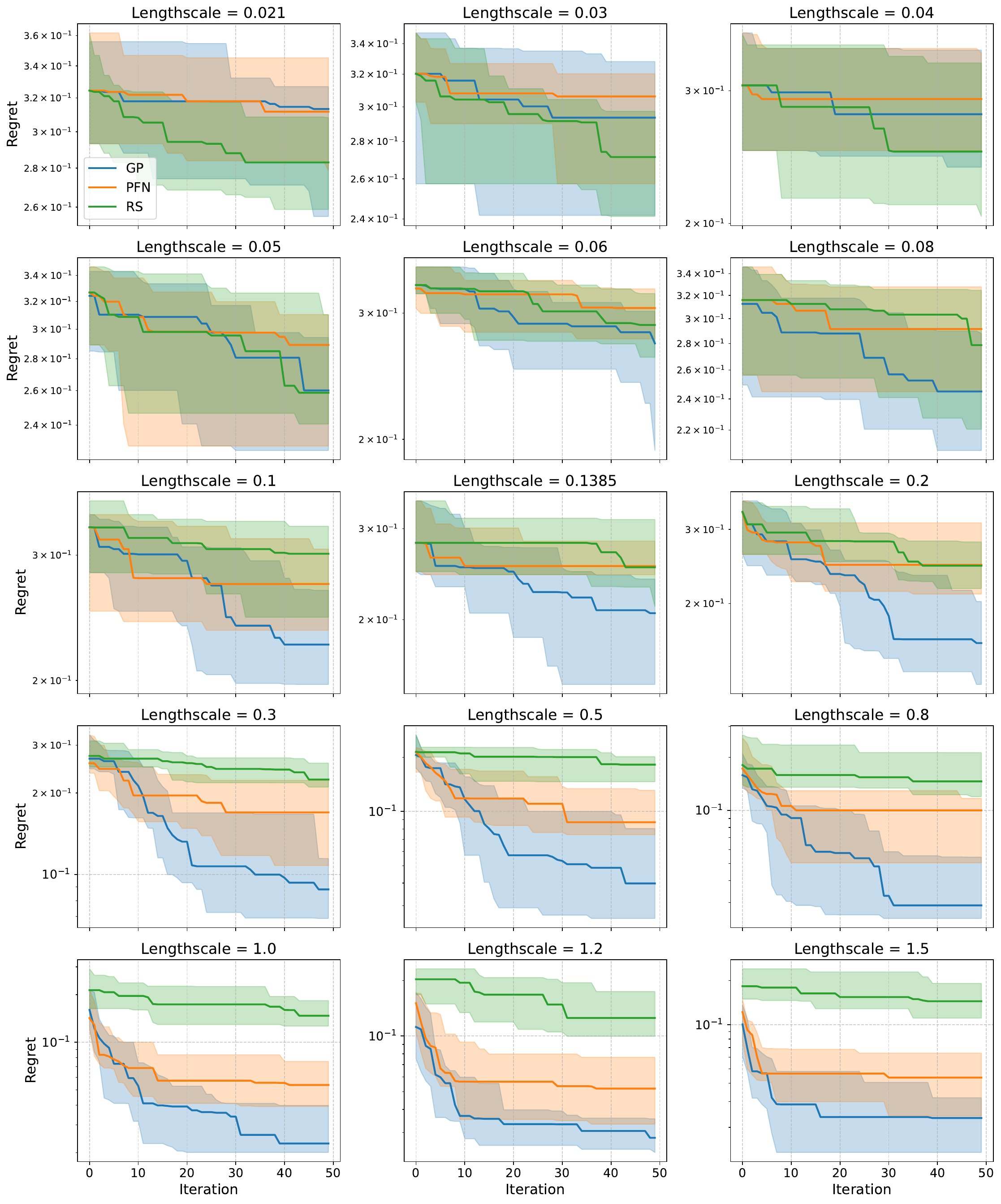}
\caption{5D normalised regret curves (median and IQR over 21 repeats) for GP, PFN and random search across 15 lengthscales. The PFN's prior mode is $l=0.1385$, panels to the left of it are increasingly rough relative to the prior, and those to the right increasingly smooth.}
\label{5D_sweep_plot}
\end{figure}

\begin{figure}[htbp]
\centering
\includegraphics[width=0.95\textwidth]{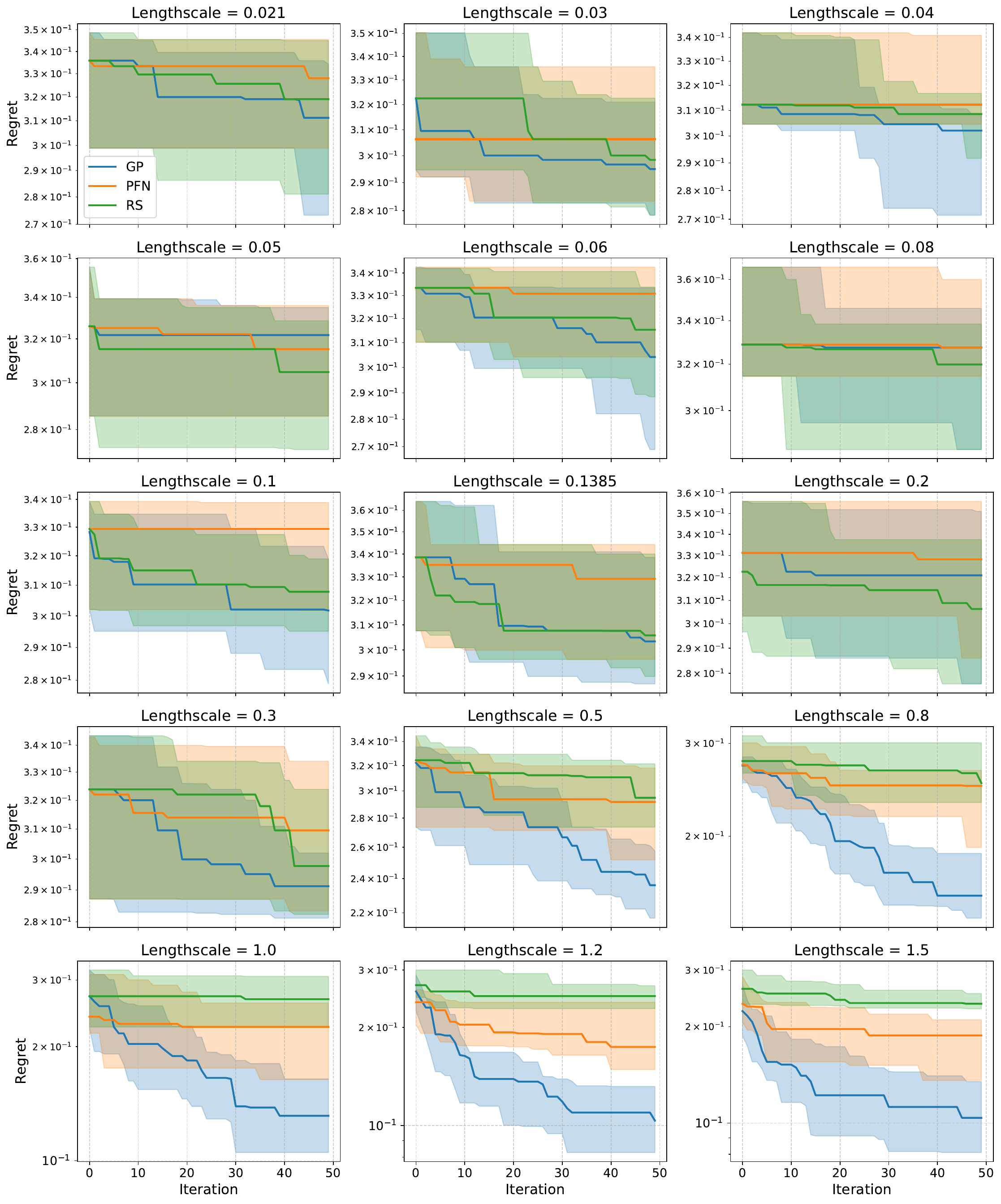}
\caption{10D normalised regret curves (median and IQR over 21 repeats) for GP, PFN and random search across 15 lengthscales. At this budget only the GP makes clear progress, and only at the longer lengthscales, consistent with the $d = 10$ onset near $l = 0.5$ in Figure~\ref{ls_linegraphs}; at the shorter ones no method separates from random search. The PFN's prior mode is $l=0.1385$, panels to the left of it are increasingly rough relative to the prior, and those to the right increasingly smooth.}
\label{10D_sweep_plot}
\end{figure}

\section{Regret Curves - ATR Comparison}
\label{app:atr-regret}
\begin{figure}[htbp]
\centering
\includegraphics[width=0.95\textwidth]{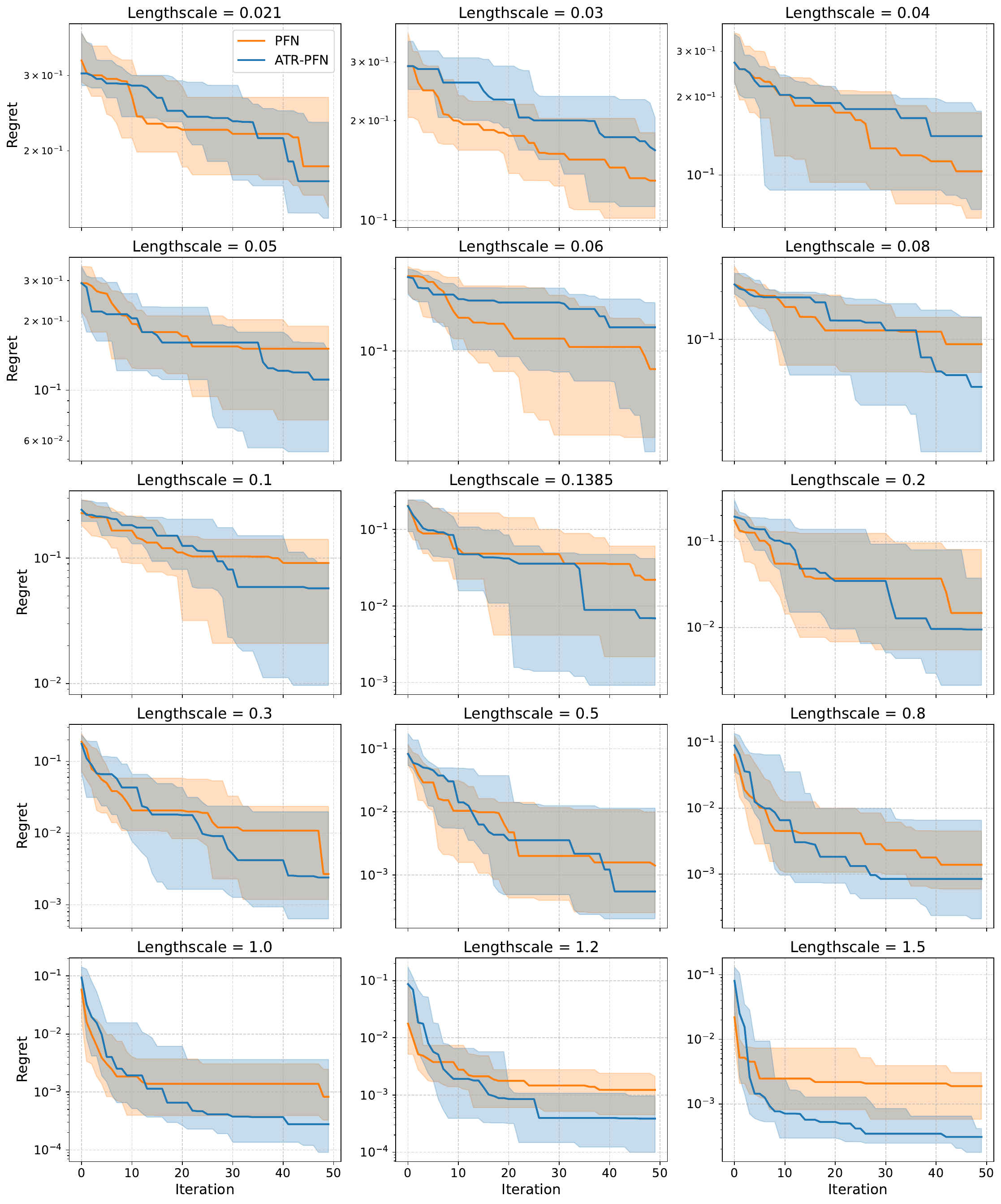}
\caption{2D normalised regret curves (median and IQR over 21 repeats) for PFN and ATR-PFN across 15 lengthscales. The PFN's prior mode is $l=0.1385$, panels to the left of it are increasingly rough relative to the prior, and those to the right increasingly smooth.}
\label{2D_sweep_plot_comp}
\end{figure}

\begin{figure}[htbp]
\centering
\includegraphics[width=0.95\textwidth]{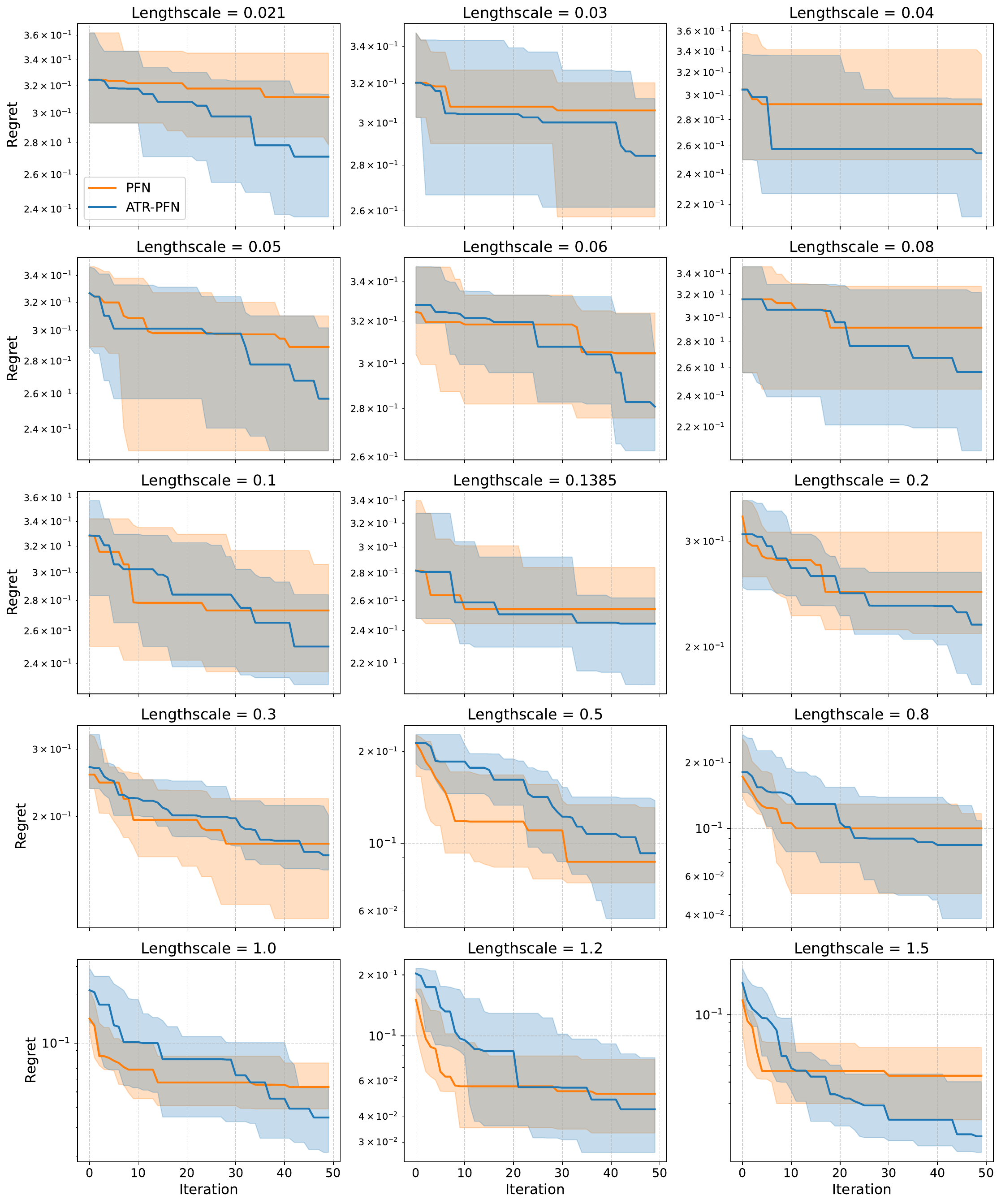}
\caption{5D normalised regret curves (median and IQR over 21 repeats) for PFN and ATR-PFN across 15 lengthscales. The PFN's prior mode is $l=0.1385$, panels to the left of it are increasingly rough relative to the prior, and those to the right increasingly smooth.}
\label{5D_sweep_plot_comp}
\end{figure}

\begin{figure}[htbp]
\centering
\includegraphics[width=0.95\textwidth]{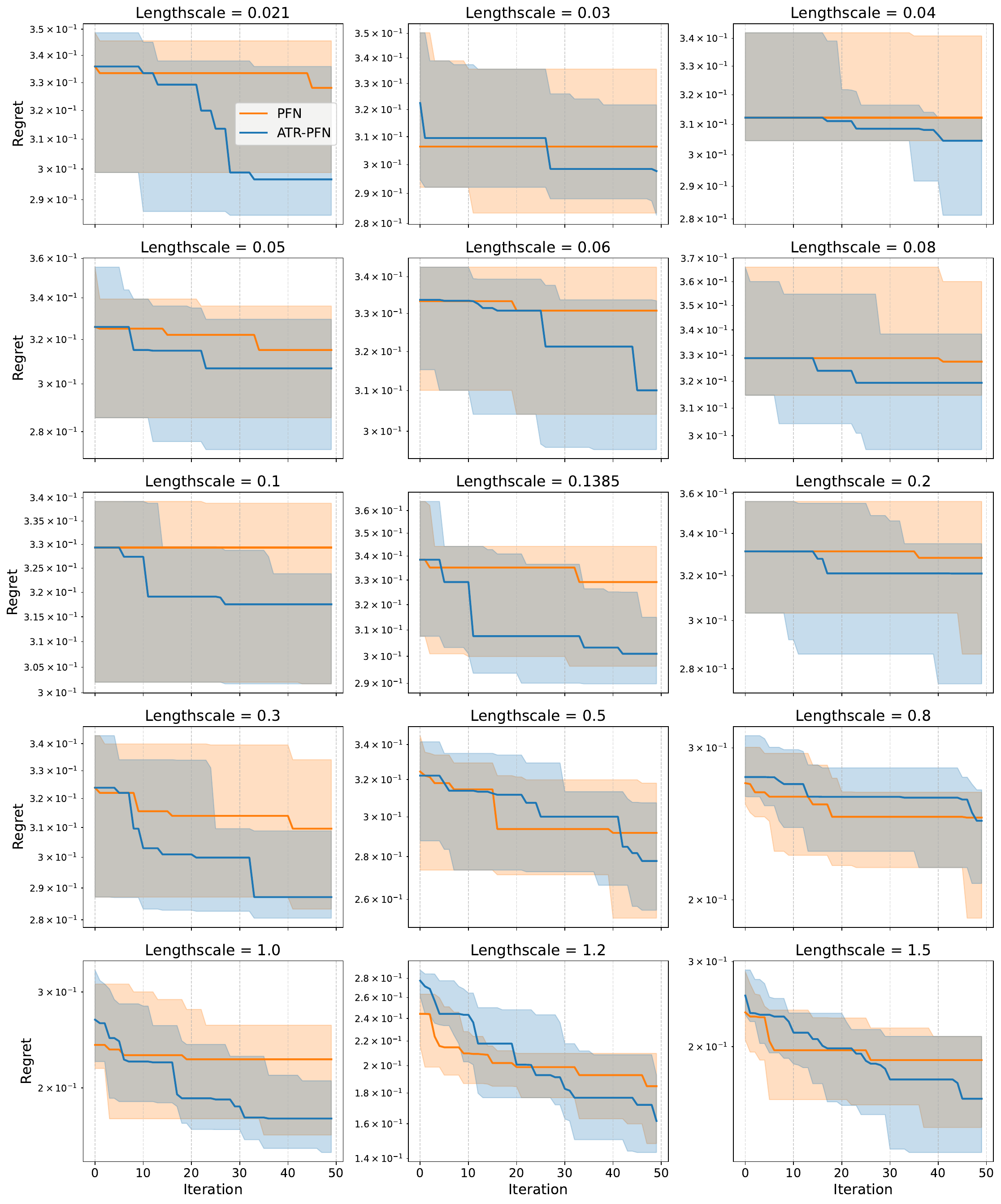}
\caption{10D normalised regret curves (median and IQR over 21 repeats) for PFN and ATR-PFN across 15 lengthscales. Both methods make limited progress at this budget up until the larger lengthscales. The PFN's prior mode is $l=0.1385$, panels to the left of it are increasingly rough relative to the prior, and those to the right increasingly smooth.}
\label{10D_sweep_plot_comp}
\end{figure}

\begin{figure}[htbp]
\centering
\includegraphics[width=0.95\textwidth]{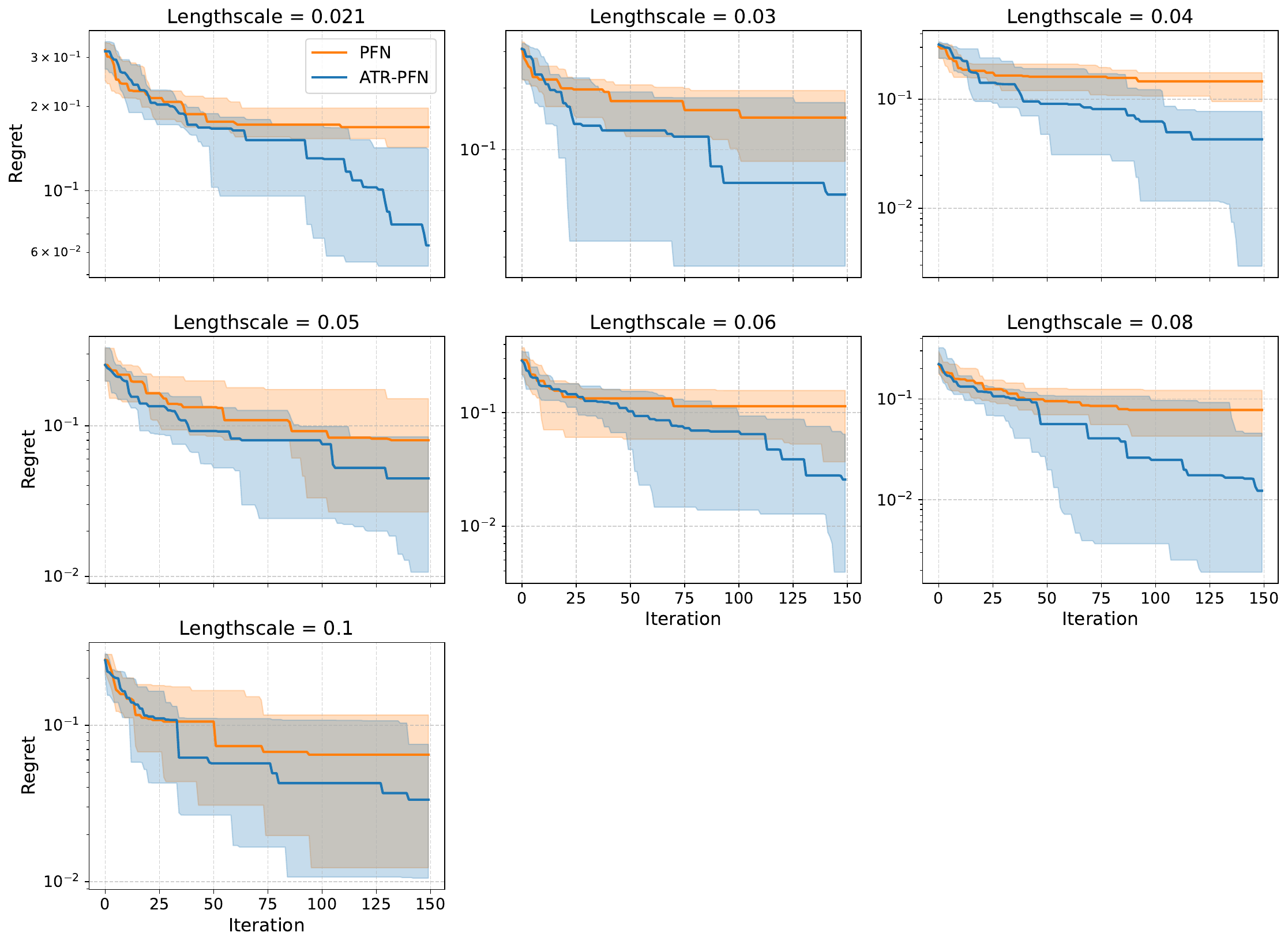}
\caption{2D normalised regret curves (median and IQR over 21 repeats) for PFN and ATR-PFN across 7 lengthscales in $ [0.021,0.1]$, run to a budget of 150 iterations rather than the 50 used elsewhere.}
\label{2D_sweep_plot_comp_150}
\end{figure}
The results in Figure~\ref{2D_sweep_plot_comp} use a budget of 50 iterations, at which the separation between PFN and ATR-PFN at the shortest lengthscale is modest. To check whether this reflects the budget rather than the method, we repeated the smaller lengthscale configurations in 2D with a budget of 150 iterations to allow sufficient information for learning at these lengthscales. The advantage widens over the longer horizon, consistent with the $k$-nearest-neighbour floor loosening as $N$ grows and permitting a tighter trust region. We do not separate
the contributions of the warp and the trust region here.

\section{Ablation Study}
\label{app:ablation}
\begin{figure}[htbp]
\centering
\includegraphics[width=0.99\textwidth]{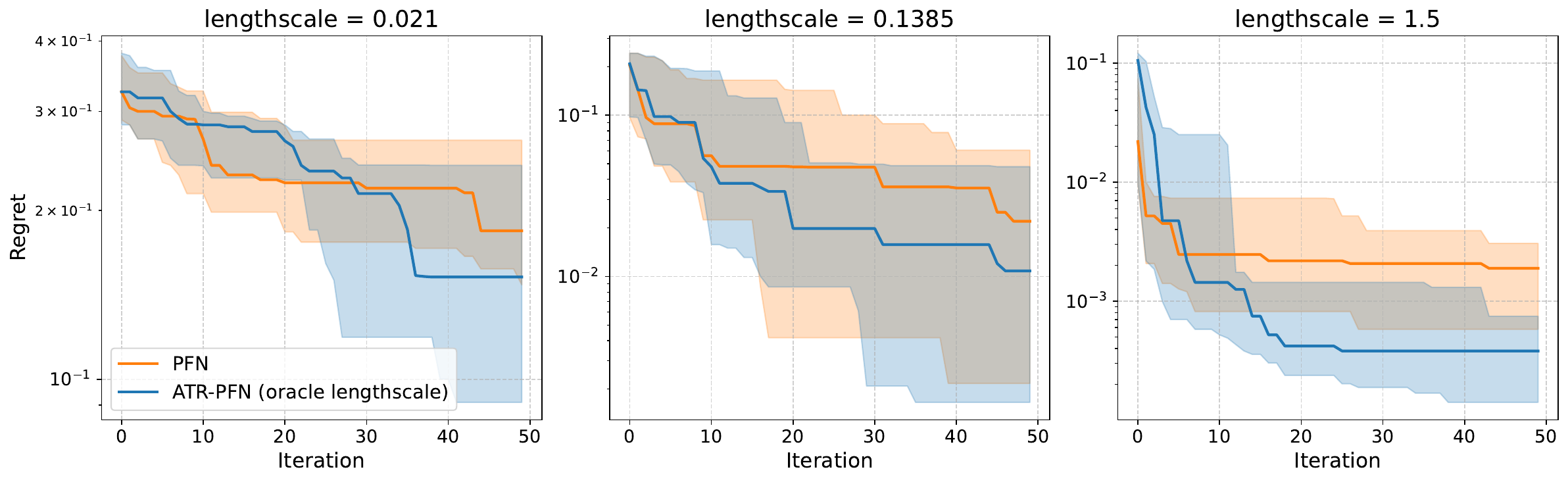}
\caption{2D ablation study. The centre panel is at the PFN's prior mode ($l=0.1385$), the left and right panels are the extremes of the tested range.}
\label{ablation_2d}
\end{figure}

We ran an ablation study to separate the effect of the warping from the estimation of the lengthscale (Figure~\ref{ablation_2d}), in 2D as it was the most representative dimension, 50 iterations being enough to gather sufficient information. Our results show that the method's improvement is not driven by estimator variance. The trust region and restart mechanisms are retained in this condition as they are integral to the method, so it isolates the estimator rather than separating the warping from localisation. The residual at $l = l_{target}$, where warping is identity, indicates that the trust region and restart mechanism also contribute to the performance improvement; above $l_{target}$ the reverse holds, since $L_j = 1$ retains all data and the full domain, leaving the improvement there attributable to the warping alone.

\section{Experiment Details}
\label{app:Exp_det}
\subsection{Gaussian Process Setup}
\label{app:Exp_det_GP}
The reference surrogate used is a Gaussian Process with a \matern-3/2 ARD kernel, to match the PFN~\citep{muller2023pfns4bo}, in $d$ dimensions, scaled by an outputscale with a Gamma(2.0, 0.15) prior~\citep{muller2023pfns4bo}. Lengthscales are given a smoothed box prior over $[0.015, 2.5]$ with $\sigma=0.01$, chosen to be uninformative over the tested range whilst remaining correctly specified. As our tested objective functions are noiseless, the GP's noise is constrained to $[10^{-5}, 10^{-3}]$ and initialised at $10^{-4}$.

\subsection{Prior-Data Fitted Network Setup}
\label{app:Exp_det_PFN}
As the PFN surrogate we use the model conditioned on the HEBO+ prior from PFNs4BO~\citep{muller2023pfns4bo} without any finetuning. Its meta-training prior places a Gamma$(\alpha = 1.211, \beta = 1.521)$ prior on the lengthscale, whose mode, $(\alpha - 1)/\beta = 0.1385$, is the value used as $l_{\mathrm{target}}$ throughout this paper. Unlike the HEBO surrogate that inspired it, this checkpoint was trained without input warping in its prior (\texttt{hebo\_warping} is disabled in the released training configuration).

\subsection{ATR-PFN Setup}
\label{app:Exp_det_ATR_PFN}
We used the following parameters for the Kernel Target Alignment optimisation process, shown in the table below. We found $k=4$ to be sufficient for our context budget (50), but this value will likely need to be reduced for smaller context sets and raised for larger context sets.
\begin{table}[htbp]
  \centering
  \begin{minipage}{0.48\linewidth}
    \centering
    \begin{tabular}{| l | r |}
      \hline
      \textbf{Parameter} & \textbf{Value} \\ \hline
      Learning Rate & 0.1 \\ \hline
      Iterations    & 100 \\ \hline
    \end{tabular}
    \caption{KTA optimiser settings.}
    \label{tab:kta}
  \end{minipage}\hfill
  \begin{minipage}{0.48\linewidth}
    \centering
    \begin{tabular}{| l | r |}
      \hline
      \textbf{Parameter} & \textbf{Value} \\ \hline
      Minimum Context Set ($k$) & 4 \\ \hline
      Restart Threshold ($\tau$) & 5 \\ \hline
    \end{tabular}
    \caption{ATR settings.}
    \label{tab:atr}
  \end{minipage}
\end{table}

Two details of the procedure are worth stating. The alignment is fit to all observations gathered so far, not only those inside the current region, so the estimate does not depend on the region it is used to construct. Targets are standardised before the alignment is computed, and the retained subset is standardised again after filtering, so that restricting the region does not present the PFN with an objective whose amplitude is inconsistent with its prior.

Cortes et al.~\citep{cortes2012algorithms} centre both kernel matrices; we centre only $K_y$. Centring $K_x$ sharpened the alignment's dependence on $l$ but raised the variance of the estimate at our sample sizes, producing over-aggressive warping which hurt optimisation performance, so we retain the uncentred spatial kernel.

\subsection{Regression Comparison Experiment}
\label{app:Exp_det_RCE}
Regression performance was assessed by drawing a function per repeat, conditioning each surrogate on $20 \cdot d$ uniformly sampled observations, and evaluating the predictive log-likelihood of the true function from the surrogate fit. We report the median across 21 repeats over 17 lengthscales in $[0.021, 2.0]$. This process was repeated for $d \in \{1, \dots, 10 \}$.

\subsection{Optimisation Experiment}
\label{app:Exp_det_OE}
Optimisation performance was assessed by drawing a function per repeat, conditioning each surrogate on $5 \cdot d$ uniformly sampled initial points. Optimisation was then run for 50 iterations, with 21 repeats over 15 lengthscales in $[0.021, 1.5]$ at $d \in \{ 2,5,10 \}$, using expected improvement (EI)~\citep{jones1998efficient} maximised over 100000 Sobol sampled candidates where the set is common to all surrogates at each step.
We report the median regret across the 21 draws, normalised by the range of the objective function, so that results are comparable across function draws and lengthscales.

\section{KTA Estimation Accuracy}
\label{app:kta}
\begin{figure}[htbp]
\centering
\includegraphics[width=0.95\textwidth]{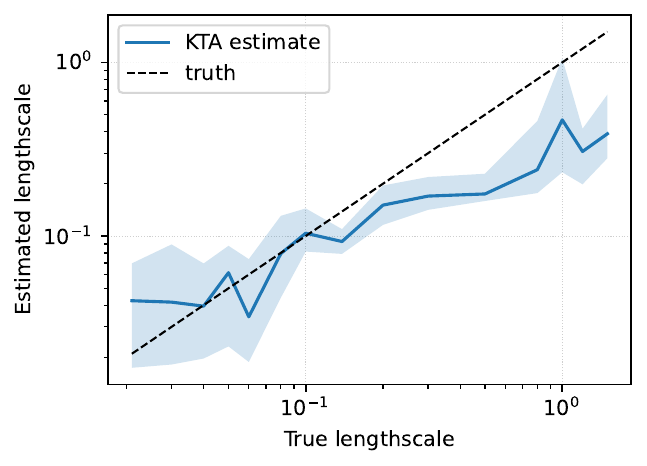}
\caption{KTA lengthscale estimates against the true lengthscale, taken at iteration 20 of the optimisation loop in 2D and averaged across the 2 axes (median and IQR over 21 repeats). The identity line is also plotted. Estimates are monotone but shrunk toward the centre of the range, so the warping factor undercorrects at both extremes.}
\label{app:kta_estimate}
\end{figure}

ATR warping depends on the lengthscale estimate being close enough to the true lengthscale for warping to move the objective function towards the prior. We assessed this in 2D over 21 repeats, taking the estimate at iteration 20 of the optimisation loop and averaging the 2 per-axis values, since the objectives were isotropic in this case. 

Shown in Figure \ref{app:kta_estimate}, the estimate is monotone in the true lengthscale but shrunk toward the centre of the tested range, crossing the identity near $l=0.1$, close to the PFN's prior mode. At both extremes the estimate causes an underestimation of the true warping factor, so the correction is partial. The estimator does not cause over-warping anywhere in the tested range, so the objective moves towards the prior rather than past it.

\section{ATR Algorithm}

\begin{algorithm2e}[htbp]
\SetAlgoLined
\caption{Anisotropic Trust Region Warping}
\label{alg:atr}
\KwData{observations $\{(x_i, y_i)\}_{i=1}^{N}$, target lengthscale $l_{\mathrm{target}}$,
        min context size $k$, restart threshold $\tau$}
\KwResult{next point to evaluate}
\While{budget remains}{
  standardise the targets, $y$\;
  estimate the lengthscale vector $l$ by maximising the alignment $A(l)$\;
  find the $k$ observations nearest $x_{\mathrm{opt}}$ in the $L_\infty$ norm with axis $j$ scaled by $l_j / l_{\mathrm{target}}$,
  and set $d_{k,j}$ to their largest deviation from $x_{\mathrm{opt}}$ in dimension $j$\;
  $s_j \leftarrow \max\!\left(2 d_{k,j},\ l_j / l_{\mathrm{target}}\right)$,
  \quad $L_j \leftarrow \min(1, s_j)$,
  \quad $u_j \leftarrow \min(1, 1/s_j)$\;
  centre the region of side $L_j$ on the anchor $x_{\mathrm{opt}}$, translating it
  inside $[0,1]^d$ if it overflows, to give bounds $[a_j, b_j]$\;
  $\mathcal{D}_{\mathcal{T}} \leftarrow \{(x_i, y_i) : x_i \in \mathcal{T}\}$\;
  $z_i \leftarrow \psi(x_i)$ for $(x_i, y_i) \in \mathcal{D}_{\mathcal{T}}$\;
  standardise the retained targets\;
  $z^\star \leftarrow \arg\max_{z \in \Pi_j [0, u_j]} \alpha\!\left(z \mid \mathcal{D}_{\mathcal{T}}\right)$\;
  $x^\star \leftarrow \psi^{-1}(z^\star)$\;
  evaluate $y^\star \leftarrow f(x^\star)$ and append $(x^\star, y^\star)$\;
  \eIf{$y^\star$ improves on the anchor}{
    $x_{\mathrm{opt}} \leftarrow x^\star$; reset failure counter\;
  }{
    increment failure counter\;
    \If{failure counter $> \tau$}{
      draw $x_{\mathrm{opt}}$ uniformly from $[0,1]^d$, evaluate it, reset counter\;
    }
  }
}
\end{algorithm2e}

\end{document}